\documentclass{article}
\usepackage{spconf,amsmath,graphicx}

\usepackage{epsfig}
\usepackage{graphicx}
\usepackage{amsmath}
\usepackage{amssymb}
\usepackage{array}
\usepackage{calc}
\usepackage{caption}
\usepackage{framed}
\usepackage{subfig}
\usepackage{enumitem}
\usepackage{tabu}
\usepackage{color,soul}
\usepackage{multirow}
\usepackage{epstopdf}
\usepackage{tabularx}

\title{Detector with Focus:\\ Normalizing Gradient in Image Pyramid}
\name{Yonghyun Kim \qquad Bong-Nam Kang \qquad Daijin Kim}

\address{Department of Computer Science and Engineering, POSTECH, Korea}

\begin{document}
%
\maketitle
\begin{abstract}
	An image pyramid can extend many object detection algorithms to solve detection on multiple scales.
	However, interpolation during the resampling process of an image pyramid causes gradient variation, which is the difference of the gradients between the original image and the scaled images.
	Our key insight is that the increased variance of gradients makes the classifiers have difficulty in correctly assigning categories.
	We prove the existence of the gradient variation by formulating the ratio of gradient expectations between an original image and scaled images, then propose a simple and novel gradient normalization method to eliminate the effect of this variation.
	The proposed normalization method reduce the variance in an image pyramid and allow the classifier to focus on a smaller coverage. 
	We show the improvement in three different visual recognition problems: pedestrian detection, pose estimation, and object detection.
	The method is generally applicable to many vision algorithms based on an image pyramid with gradients.
\end{abstract}

\section{Introduction}
\label{sec:intro}
Gradient and image pyramid are one of the essential parts for computer vision.
Well-known methods based on magnitudes and orientations of gradients are Histogram of Oriented Gradients (HOG)~\cite{dalal2005histograms}, Scale-Invariant Feature Transform (SIFT) keypoint~\cite{lowe1999object}, and Aggregated Channel Feature (ACF)~\cite{dollar2014fast}.
An image pyramid~\cite{adelson1984pyramid} is a collection of resampled images from an original image; the pyramid is used to make a computer vision problem invariant over multiple scales.
Many object detectors (e.g., {ACF-AdaBoost}~\cite{dollar2014fast}, and {Viola and Jones}~\cite{viola2004robust}), scan a detection window of a fixed size over an image pyramid.

However, interpolation while constructing the image pyramid usually causes a difference between the gradients of the original image and the scaled image~\cite{ruderman1994statistics}.
When pixels in downsampled images are computed using a bilinear function over corresponding pixels, the intensities of the pixels have similar distribution and magnitude, but the skipped pixels in downsampled images increase gradients (the first derivative of intensity).
In contrast, the inserted pixels in upsampled images decrease the gradients.
We define this difference between original image and scaled image as gradient variation.

\begin{figure*}[t]
	\centering
	\fbox{\includegraphics[width=17.3cm]{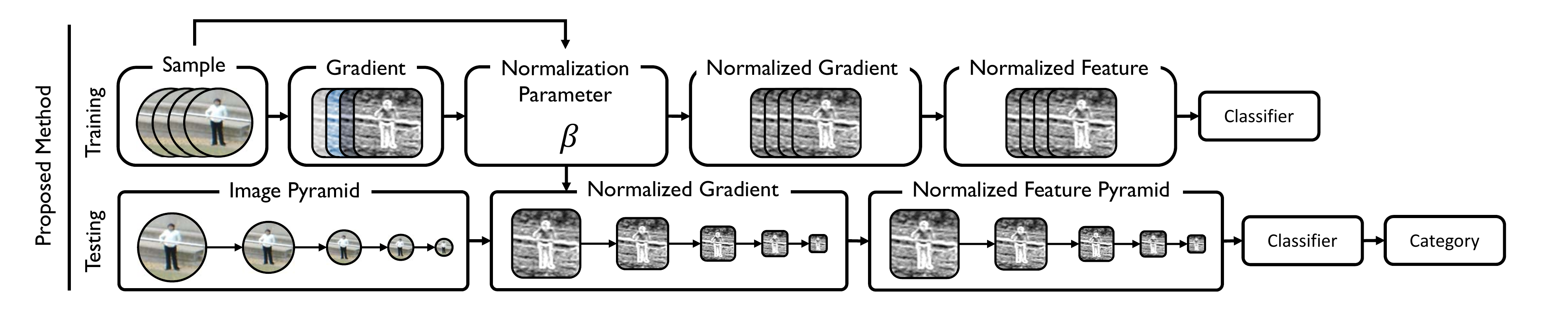}}
	\caption{
		The proposed method constructs an image pyramid, and computes normalized gradients using the proposed normalization function.
		Unlike an image pyramid and a fast feature pyramid, the proposed method enhances the quality of samples in both training and testing to improve the accuracy of classifiers.
	}
	\label{fig:comp}
\end{figure*}

Our method is inspired by the gradient variation that causes the decrease in accuracy of the classifiers.
The increased variance of gradients over the image pyramid increases the coverage of the classifier.
Thus, the increased coverage decreases the accuracy and precision of the classifiers~\cite{geman1992neural,jones2003fast,park2010multiresolution,yan2013robust}.

Hence, we propose a simple and novel gradient normalization method by analyzing the gradient variation in the viewpoint of the classifier (Fig.~\ref{fig:comp}).
The proposed method defines the original image as reference, and normalizes gradients from other resampled images to the reference image.
The normalized gradient, which is similar to the gradients of original images, reduces the variance, and increases the performance of the classifiers with negligible increase in computing time.

\vspace*{-0.2cm}
\section{Gradient Variation in Multi-Scale}

In this section, we discuss the change of gradients that occurs in an image pyramid, which is used to apply the fixed-size detector to multi-scale detection.

\subsection{Analysis of Gradient in Multi-Scale}

We compared the difference between original images and scaled images that include objects of the same identity, and observed that the first derivatives of intensity are greater in downsampled images than in the original images, even if the distributions of intensity are similar~\cite{dollar2014fast,ruderman1994statistics,dollar2010fastest}.

We theoretically show the gradient difference by computing the ratio of gradient expectations between the two images under three conditions: computing gradient using a central difference method~\cite{gonzalez2009digital}, sampling images using a bilinear interpolation~\cite{gonzalez2009digital}, and decomposing the problem to a one-dimensional form.


Let an image $f_s$ be a sequence that consists of the pixels~$f_s(x)$ from an upsampled image with scale $s$. 
$f_{r=1.0}$ is a reference image, which is original and the only natural data in an image pyramid.
A linear interpolation computes an upsampled image $f_{s}$ by inserting $z$ new pixels between two adjacent pixels on the original image. 
The pixels on the upsampled image are partitioned into inherited pixels and interpolated pixels.
The number of inserted pixels between two adjacent inherited pixels is $z=s-1$ and is therefore an integer $\geq$ 0.
The pixels in an upsampled image consist of a set of inherited pixels and $z$ sets of inserted pixels, so the total number of pixels is $n_s = (n_r-1)z+n_r$.
The pixels in an upsampled image is approximated as

\begin{equation}
\begin{aligned}
f_s(x)=
\frac{(z+1-d)f_{r}(x) + df_{r}(x+1)}{z+1},
\end{aligned}
\end{equation}
where $d$ is the distance between $x$ and the nearest inherited pixel leftward.

We use a central difference function~${\phi(f_s)}$ as a gradient function and an intermediate difference function~$\tilde{\phi}(f_s)$ is appeared in the calculation of gradient expectation, by substituting. 
When a gradient is computed at $x$, $\tilde{\phi}(f_s)$ subtracts the pixel at $x$ from the adjacent pixel at $x+1$, and ${\phi(f_s)}$ subtracts the neighbor pixels at $x-1$ and $x+1$:  

\begin{equation}
	\begin{aligned}
		\tilde{\phi}(f_{s}) = \left [ \frac{\partial f_{s} }{\partial x} \right ] _ {\Delta = 1} &\approx \left| f_{s}(x+1) - f_{s}(x) \right| \\
		\phi(f_{s}) = \left [ \frac{\partial f_{s} }{\partial x} \right ] _ {\Delta = 2} &\approx \left| f_{s}(x+1) - f_{s}(x-1) \right| ,\\
	\end{aligned}
\end{equation}
where $\Delta$ is an interval of a differential.

To prove the existence of the gradient difference, we compute the gradient expectation~$E[\phi(f_s)]$ at scale $s$:
\begin{equation}
\small
\begin{aligned}
E\left[\phi(f_{s})\right]=&
\frac{1}{(z+1)n_r-z-2}
\left\lbrace
\sum_{x=2}^{n_r-1}\frac{\left| f_{r}(x+1)-f_{r}(x-1) \right| }{z+1}
\right.
\\
& \hspace{24mm}  + \left.
\sum_{x=1}^{n_r-1}\frac{2z\left| f_{r}(x+1)-f_{r}(x) \right| }{z+1}
\right\rbrace
\\
=&
\frac{
	(n_r-2)E\left[\phi(f_{r})\right]
	+
	(2z(n_r-1))E\left[\tilde{\phi}(f_{r})\right]
}{(z+1)((z+1)n_r-z-2)}.
\end{aligned}
\end{equation}

The input images that are used in object detection typically have enough pixels to assume that $n_r$ is infinite.
$E\left[\phi(f_{s})\right]$ is approximated as $\lim_{n_r \rightarrow \infty} E\left[\phi(f_{s})\right]$:

\begin{equation}
\begin{aligned}
E\left[\phi(f_{s})\right] & \approx 
\lim_{n_r \rightarrow \infty} E\left[\phi(f_{s})\right]
\\
&=
\frac{E\left[ \phi(f_r) \right]+2(s-1) E\left[\tilde{\phi}(f_{r})\right]}{s^2}.
\end{aligned}
\label{eq:egR}
\end{equation}

Eq.~\ref{eq:egR} reveals that a gradient difference between the upsampled and reference image exists, and is determined by scale and the gradient expectation of an intermediate difference function~$\tilde{\phi}(f_s)$.

\vspace*{-0.2cm}
\subsection{Formulation of Gradient Variation}

We define gradient variation as the difference of gradient expectation between an original image and a scaled image, and formulate the variation as the ratio of the gradient expectations.
We formulate the equations for the integer variable~$z$, however, the practical algorithm estimates the real value of $z$ through nearest neighbor or linear approximation.
With the same concept, we expand the equations of the gradient variation to a real value.
Gradient variation~$\rho(f_s|f_r)$ between the upsampled image~$f_s$ and the reference image $f_r$ is computed as

\begin{equation}
	\begin{aligned}
		\rho(f_s|f_r)=\frac{E\left[\phi(f_{s})\right]}{E\left[\phi(f_{r})\right]}
		=\frac{2cs-2c+1}{s^2}
	\end{aligned}
	\label{eq:gm}
\end{equation}
where $c={E[\tilde{\phi}(f_{r})]}/{E[\phi(f_{r})]}$ is a constant. 
Because Eq.~\ref{eq:gm} is only available for upsampled images due to the definition of $f_s$, we replace the reference image and the scaled image with each other to represent downsampling.
We invert $s$ to re-define it to the range $(0, 1]$, then calculate the inverse of $\rho(f_{s}|f_r)$ as

\begin{equation}
\overline{\rho}(f_s|f_r)=
\frac{1}{\rho(f_{1/s}|f_{r})}=\frac{1}{(1-2c)s^2+2cs}.
\end{equation}

The practical interval $[1, 2)$ of $s$ for upsampling has a smaller rate of change than the interval $(0, 1]$ of $s$ for downsampling, and the constant $c$ is close enough to $0.5$ for degree reduction ($\mu_c=0.62$ and $\sigma_c=0.05$ in INRIA dataset).
We approximate the last term as~$1-2c \approx 0$ in the numerator for upsampling, to simplify the gradient variation~$\rho(f_{s}|f_r)$.

The gradient variation for resampled images are computed as

\begin{equation}
	\begin{aligned}
		&\rho(f_{s}|f_r)
		\\
		&=\begin{cases}
			\rho(f_{s}|f_r) \approx (2c/s) & 1<s \\
			\overline{\rho}(f_{s}|f_r)=1/\{(1-2c)s^2+2cs\} &  0<s \leq 1
		\end{cases}.
		\label{eq:comb1}
	\end{aligned}
\end{equation}

Eq.~\ref{eq:comb1} shows that $\rho(f_{s}|f_r)$ is a decreasing function.
These trends imply that upsampling decreases gradients and downsampling increases gradients.
This phenomenon implies that the gradient distribution of the resampled images is different from the gradient distribution of the reference images; the increased variance increases the difficulty of training the classifiers~\cite{geman1992neural,james2013introduction}.

\vspace*{-0.2cm}
\section{Gradient Normalization}

We propose a normalization method to eliminate the gradient variation.
The proposed method normalizes the gradients of the resampled image to the gradients of the reference image to reduce the variance of gradients.
The reduced variance makes the classifier concentrate on a small coverage, and improves overall precision and accuracy of detection~\cite{geman1992neural,jones2003fast,park2010multiresolution,yan2013robust}.
We obtain the gradient normalization function $g(s)$ as the inverse of the gradient variation~$\rho(f_{s}|f_r)$ as
\begin{equation}
	\begin{aligned}
		g(s) &= \frac{1}{\rho(f_{s}|f_r)}
		\\
		&=\begin{cases}
			{s}/(2c) \approx a_1s+b_1 & 1<s \\
			(1-2c)s^2+2cs \approx a_2s^2+b_2s+c_2 &  0<s \leq 1
		\end{cases}.
	\end{aligned}
\end{equation}

with a bias term: $b_1$ and $c_2$.

The normalization function consists of polynomials of degree 1 for upsampling and of degree 2 for downsampling. 
We compute the optimal coefficients of $g(s)$ for the training set.
Given a training image~$f^k$, we define an error criterion~$\mathcal{E}$, which is a mean squared error to minimize the difference between the normalized gradient and the reference gradient:

\begin{equation}
	\mathcal{E}=
	\sum_{s \in S} 
	{
		\sum_{k \in K} 
		{\left [ \phi ( f_{s}^k )g(s) - \phi ( f_{r}^k ) \right ]}
	} ^ 2, 
	\label{eq:ls}
\end{equation}
where $S$ is a set of scales and $K$ is a set of training images.

The separate training of the normalization functions~$g(s)$ for upsampling and downsampling requires an equality constraint. 
We impose an equality constraint between original and normalized gradients at the reference scale.
The equality constraint prevents gradient normalization at reference images and keeps the continuity of the gradient normalization function at reference image, and is defined as
\begin{equation}
	\sum_{k \in K} \left[ \phi ( f_{r}^k )g(r) - \phi ( f_{r}^k ) \right] = 0. 
	\label{eq:eqc}
\end{equation}

The error criterion and the equality constraint is combined into a Lagrangian
\begin{equation}
	\begin{aligned}
		&L(\beta_d,\beta_u,\lambda_d,\lambda_u)
		\\
		&=\begin{cases}
			\left \| X_{d}\beta_d-y_d \right \|^2+\lambda_d(X_r\beta_d-y_r), & 1<s \\
			\left \| X_{u}\beta_u-y_u \right \|^2+\lambda_u(X_r\beta_u-y_r), & 0<s \leq 1 \\
		\end{cases}
		,
		\label{eq:lag}
	\end{aligned}
\end{equation}
where subscripts $i = d, u$ and $r$ represent downsampled, upsampled and reference, respectively, $X_i$ are Vandermonde matrices of scales, $\beta_i$ are coefficients of the proposed polynomial equation, $y_i$ are vectors of the ratio of gradients, and $\lambda_i$ are Lagrange multipliers.
The optimal coefficients $\beta^*$ of $g(s)$ are computed by minimizing the Lagrangian~\cite{duda2012pattern}.

\begin{figure}[t]
	\centering
	\fbox{\includegraphics[width=8cm]{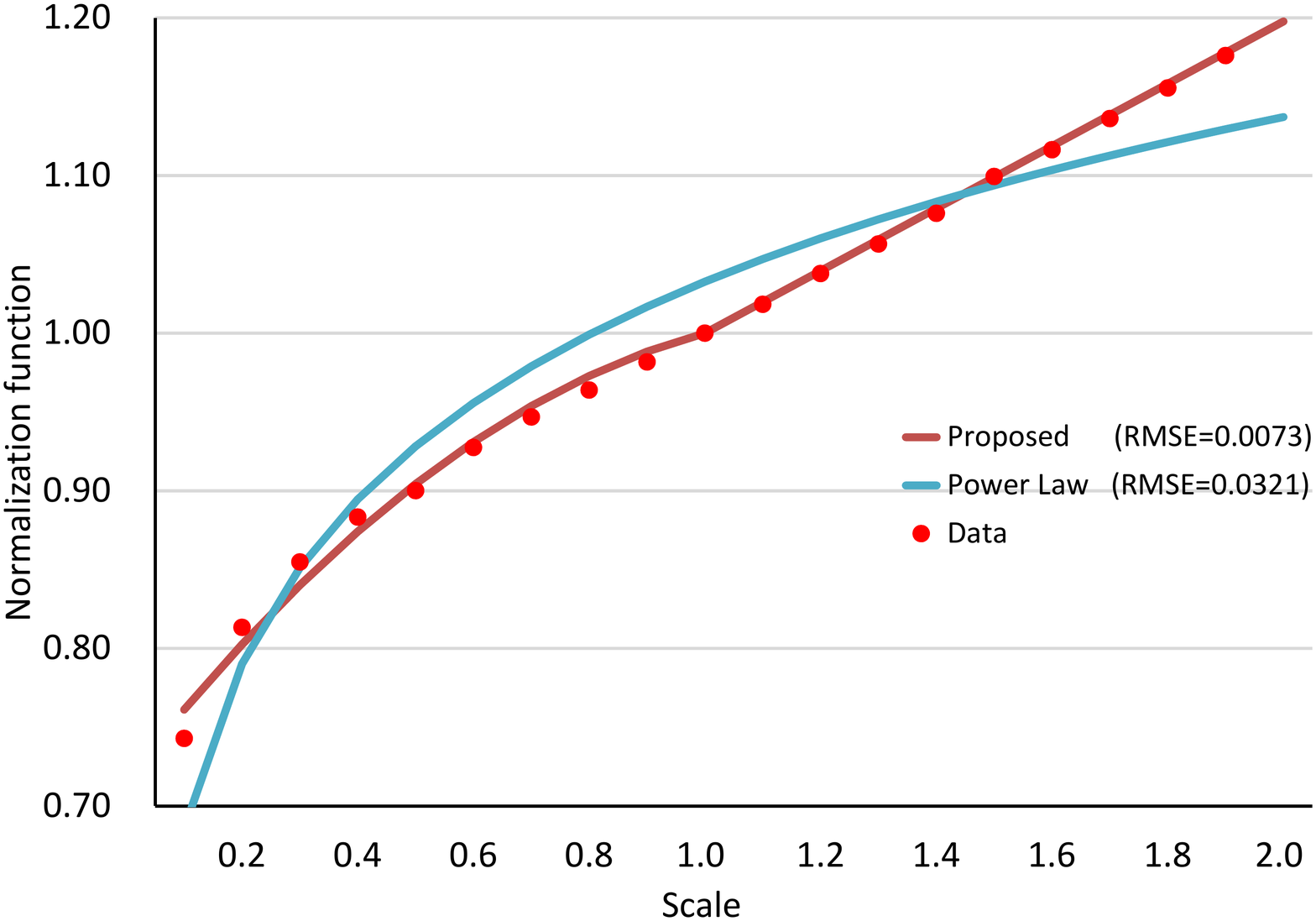}}
	\caption{Illustration on the collected data for the normalization function from scale $0.1$ to scale $2.0$, and on the estimated value using the proposed function and a power law. 
		Our normalization function fits the data over every scales, whereas the power law fails the extremes; our function also has smaller RMSE~($0.0073$) than RMSE~($0.0321$) of the power law.
		The data is collected in INRIA dataset.
	}
	\label{fig:all1}
\end{figure}

We compared the fitting accuracy of the gradient normalization between the proposed function and a power law function (Fig.~\ref{fig:all1}).
A power law was dealt with to represent the study of natural image statistics by Ruderman and Bialek~\cite{ruderman1994statistics} and Dollar et al.~\cite{dollar2014fast}.

\section{Experiments}

We show the effectiveness of the gradient normalization in object detection with three applications: pedestrian detection, pose estimation, and object detection.

\vspace*{-0.4cm}

\subsection{Pedestrian Detection}
\label{sec:ped}

\begin{figure}[t]
	\centering
	\subfloat{
		\fbox{\includegraphics[width=8cm]{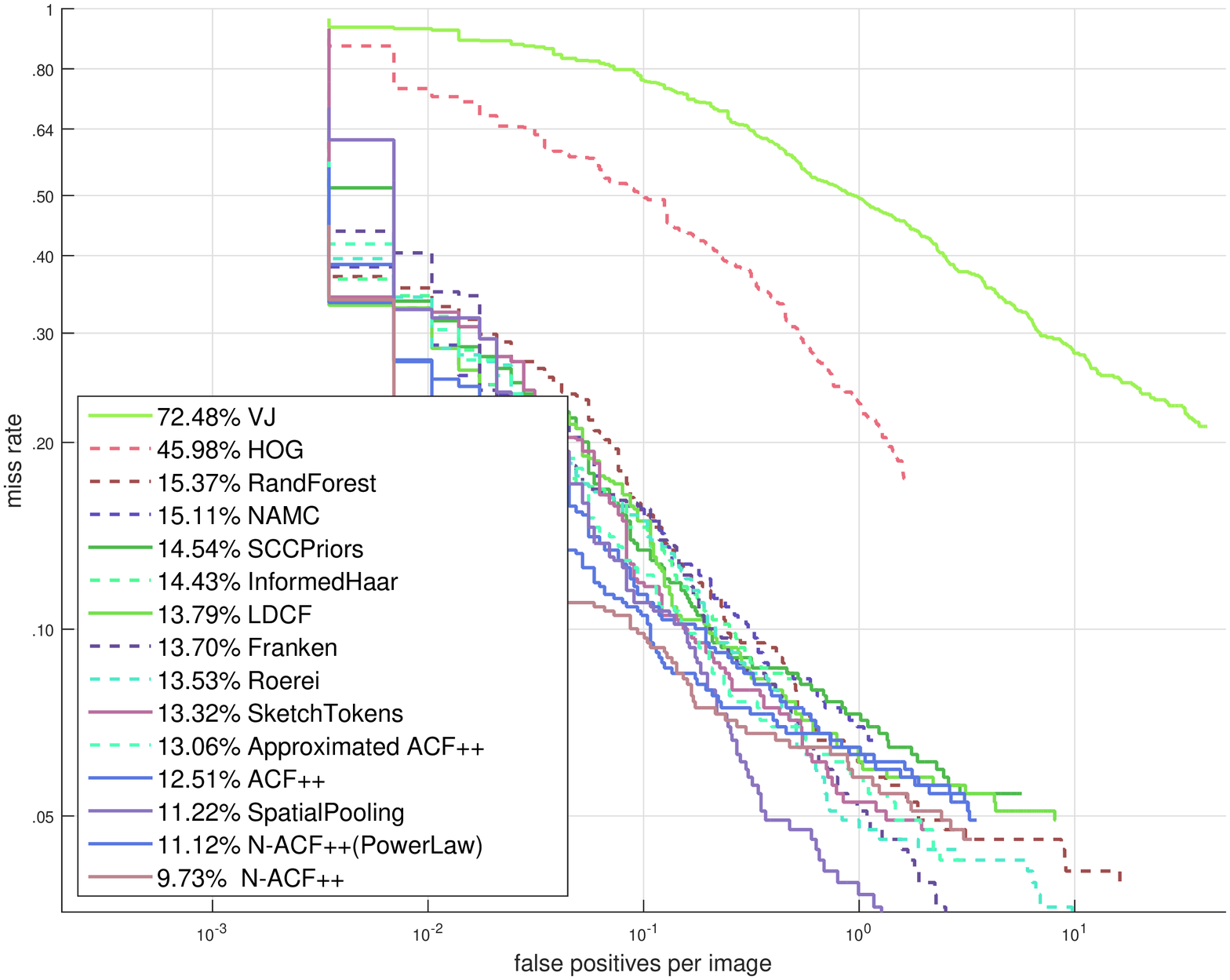}}
	}
	\caption{The log-average miss rate of \textbf{ACF++}, \textbf{Approximated ACF++}, \textbf{N-ACF++(PowerLaw)}, \textbf{N-ACF++} on INRIA dataset.}
	\label{fig:inriaPerf}
\end{figure}

\textbf{ACF}~\cite{dollar2014fast,acfcode} is widely used for pedestrian detection~\cite{nam2014local,zhang2015filtered}.
In this paper, we build \textbf{ACF++}, which is a simplified version of the filtered channel features based detector~\cite{zhang2015filtered}.
We combine the original ACF and the differences between two neighboring features, which are part of the checkerboards filters.
\textbf{Approximated ACF++} is a version of \textbf{ACF++} with a fast feature pyramid.
We evaluate ACF++ using normalized gradient (\textbf{N-ACF++}) on INRIA dataset~\cite{dalal2005histograms}.
\textbf{N-ACF++} is trained in the same way as \textbf{ACF++} without gradient normalization.
To train the gradient normalization function, we collected all gradient expectations of both positive and negative images over scales from $0.1$ to $2.0$ in increments of $0.1$.
We applied our normalization method in both training and testing, and we only normalized gradient magnitudes to naturally spread out over the gradient-based features such as HOG.
\textbf{N-ACF++(PowerLaw)} is a version of \textbf{N-ACF++} trained by a power law function.
The proposed normalization method with \textbf{ACF++} shows the improvement from 12.51\% to 9.73\% log average miss rate (Fig.\ref{fig:inriaPerf}).

\vspace*{-0.4cm}

\subsection{Pose Estimation}
\label{sec:pose}
Yang et al.~\cite{yang2013articulated,articulated} proposed flexible mixtures of parts model (\textbf{FMM}) to estimate human poses.
Each appearance model is trained as a filter of HOG~\cite{dalal2005histograms} based features that consist of contrast-sensitive HOG, contrast-insensitive HOG, and magnitudes.
We evaluate the normalized \textbf{FMM} (\textbf{N-FMM}) on PARSE dataset~\cite{ramanan2006learning}.
As the negative images, we used the INRIA dataset~\cite{dalal2005histograms}.
We achieved 2\%p overall improvement on probability of correct keypoint (Table~\ref{tb:parse}).

\begin{table}[!h]
	\hspace{-0.2cm}
	\scriptsize
	\begin{tabu}{l|[1pt]c|c|c|c|c|c|c|c}
		& Avg & Head & Shou & Elbo & Wris & Hip & Knee & Ankle \\ 
		\tabucline[1pt]{-}
		\textbf{FMM}~\cite{yang2013articulated} & 72.3 & 89.0 & 85.3 & 66.0 & 46.3 & 76.5 & 76.3 & 66.3 \\ 
		\hline \textbf{N-FMM} & {74.2} & {91.0} & {86.8} & {67.6} & {49.5} & {80.2} & {77.6} & {66.8} 
	\end{tabu} 
	\caption{Probability of correct keypoint for $\textbf{FMM}$ and \textbf{N-FMM} (using normalized gradients) on PARSE dataset.}
	\label{tb:parse}
\end{table}

\subsection{Object Detection}
\label{sec:obj}

The deformable part model (\textbf{DPM}) from Felzenszwalb et al.~\cite{felzenszwalb2010object,voc-release5} is a representative approach for object detection.
\textbf{DPM} consists of mixtures of multiscale deformable part models that are trained using partially labeled data, and each part model includes appearance and spatial models.
Appearance models are trained as a filter of HOG~\cite{dalal2005histograms} based features that consist of contrast-sensitive HOG, contrast-insensitive HOG, and magnitudes.
We evaluate the normalized \textbf{DPM} (\textbf{N-DPM}) on PASCAL 2007 dataset~\cite{pascal-voc-2007}.
We achieve 1\%p overall improvement and 4.4\%p maximum improvement in average precision scores~\cite{pascal-voc-2007} (Table~\ref{tb:voc}).

\begin{table}[!h]
	\centering
	\scriptsize
	\begin{tabu}{>{\centering}m{1cm}|[1pt]>{\centering}m{1cm}|>{\centering}m{1cm}|>{\centering}m{1cm}|[1pt]>{\centering}m{1cm}|>{\centering}m{1cm}}
		& DPM & \hspace{-1.4mm}N-DPM &        & DPM & \hspace{-1.4mm}N-DPM \\
		\tabucline[1pt]{-} 
		plane  & 33.3 & 34.2 			   & table  & 24.6    & \textbf{27.3}      \\
		bike   & 59.7 & \textbf{60.7}      & dog    & 12.2    & 12.5      		   \\
		bird   & 10.4 & 10.8      		   & horse  & 56.4    & 57.0      		   \\
		boat   & 15.5 & \textbf{16.6}      & mbike  & 47.7    & \textbf{48.9}      \\
		bottle & 27.1 & 27.2      		   & person & 42.6    & 43.2      		   \\
		bus    & 51.2 & \textbf{52.8}      & plant  & 14.3    & 14.5      		   \\
		car    & 58.2 & 58.2      		   & sheep  & 18.6    & \textbf{23.0}      \\
		cat    & 23.9 & \textbf{25.5}      & sofa   & 37.6    & 37.8		       \\
		chair  & 19.9 & \textbf{21.3}      & train  & 45.5    & \textbf{46.8}      \\
		cow    & 25.1 & 25.7               & tv     & 43.4    & 43.5     
	\end{tabu}
	\caption{Average precision scores for \textbf{DPM} and \textbf{N-DPM}~(using normalized gradients) on PASCAL VOC 2007.}
	\label{tb:voc}
\end{table}

\vspace*{-0.6cm}
\section{Conclusion}

Our research reinterprets the gradient variation in the viewpoint of the classifier.
Unlike conventional approaches concentrating on computing resized images, our approach concentrates on decreasing the coverage of the classifier to enhance the focus of the classifier.
We prove the existence of the gradient variation by formulating the ratio of gradient expectations between an original image and scaled images, then estimate a normalization function to eliminate the effect of this variation.
Our calculations and experiments prove the validity of the gradient normalization function.
The proposed method is not restricted to object-detection applications, but can be applied in many gradient-based studies with negligible cost of computing time.
We will adopt our study to deep learning based features.

\section{ACKNOWLEDGEMENT}

This work was supported by Institute for Information \& communications Technology Promotion (IITP) grant funded by the Korea government (MSIP)(2014-0-00059, Development of Predictive Visual Intelligence Technology), MSIP (Ministry of Science, ICT and Future Planning), Korea, under the “ICT Consilience Creative Program” (IITP-R0346-16-1007) supervised by the IITP, and MSIP(Ministry of Science, ICT and Future Planning), Korea, under the ITRC (Information Technology Research Center) support program (IITP-2017-2016-0-00464) supervised by the IITP.																		

\bibliographystyle{IEEEbib}
\bibliography{egbib}

\end{document}